\documentclass[11pt,a4paper]{article}
\usepackage[a4paper, total={7in, 9.5in}]{geometry}
\usepackage[latin1]{inputenc}
\usepackage{amsmath}
\usepackage{amsfonts}
\usepackage{amssymb}
\usepackage{graphicx}
\usepackage{bm,subfig,url,balance}
\usepackage[colorlinks=true]{hyperref}

\title{Performance evaluation of deep neural networks for forecasting time-series with multiple structural breaks and high volatility}

\author{Rohit Kaushik, Shikhar Jain, Siddhant Jain, Tirtharaj Dash}

\date{%
	\small{\em Department of Computer Science and Information Systems\\
		Birla Institute of Technology and Science Pilani\\
		K.K. Birla Goa Campus, Zuarinagar, Goa 403726, India\\
		Email: \{rohit.koush, jain.shikhar97, jainsiddhant29\}@gmail.com, dashtirtharaj@acm.org}
}

\begin{document}
	
\maketitle

\begin{abstract}
The problem of automatic and accurate forecasting of time-series data has always been an interesting challenge for the machine learning and forecasting community. A majority of the real-world time-series problems have non-stationary characteristics that make the understanding of trend and seasonality difficult. Our interest in this paper is to study the applicability of the popular deep neural networks (DNN) as function approximators for non-stationary TSF. We evaluate the following DNN models: Multi-layer Perceptron (MLP), Convolutional Neural Network (CNN), and RNN with Long-Short Term Memory (LSTM-RNN) and RNN with Gated-Recurrent Unit (GRU-RNN). These DNN methods have been evaluated over 10 popular Indian financial stocks data. Further, the performance evaluation of these DNNs has been carried out in multiple independent runs
for two settings of forecasting: (1) single-step forecasting, and (2) multi-step forecasting. These DNN methods show convincing performance for single-step forecasting (one-day ahead forecast). For the multi-step forecasting (multiple days ahead forecast), we have evaluated the methods for different forecast periods. The performance of these methods demonstrates that long forecast periods have an adverse effect on performance.
\end{abstract}

\section{Introduction}
\label{sec:intro}

In recent years, with the development of countries, the stock market is becoming a more and more essential and intricate part of their economy. One such study can be found in~\cite{mishra2015random}. Nowadays, investors investing in stocks need to consider a large number of factors and evaluate a considerable amount of risks before investing in it in any form~\cite{lehkonen2015democracy}. This issue is because of the chaotic and dynamic nature of the stock prices in the present times. These investors expect to make decent profits after the investments. However, analysing factors and risks affecting the stock prices and predicting them could be highly exhaustive. They could require a higher degree of skilled task~\cite{dimic2015political}. Hence, the prediction of stock prices could be a significant reference for the investors and financial pundits for trading and investing strategies.

With the streaming developments in machine learning (ML) tools and techniques, especially deep learning (DL) algorithms along with an adequate increase in the potential of computational power, predicting stock prices have become less hectic and does not require much skill on the economic fronts. These DL tools and algorithms, such as Deep Neural Networks (DNNs), would learn the trend and factors responsible for the fluctuations (like sudden rise or drop) in the prices and accordingly predict values with acceptable approximations~\cite{cavalcante2016computational}. Furthermore, the primary advantage of such methods is that they may be able to handle the raw time-series suitably and forecast the future raw outputs. These outputs, however, could be one or multiple: respectively, we can call it as `single-step' and `multi-step' forecasting.

Recently, there have been many successful attempts to use machine learning methods for automatic time-series forecasting. Some of these methods do incorporate the information from social media, some ways deal with a transformed feature space, and some work with various economic indicators. One could follow some recent works that are published under this umbrella in~\cite{schoen2013power,su2016hybrid,tsai2018forecasting,panigrahi2020study}.

In this paper, we employ and explore various state-of-the-art deep neural network methods to build models predicting stock prices. As we wish the model to analyse and understand the factors affecting the prices over a time period and predict accurately, this problem could also be treated as a kind of time-series analysis problem, where the goal is not only to predict the stock prices but instead show some understanding of the effects of volatility and structural breaks on the prediction~\cite{stoll1988volatility,choi2010long}. In what follows, we outline our significant objectives and contributions to this work.

\subsection{Objectives and contributions of the study}
Our goal is to study the performance of neural machine learning models towards forecasting the prices of stocks that have exhibited a significant degree of volatility with numerous structural breaks. Our study is focused on the application of deep neural networks. To the best of our knowledge, less number of studies has been conducted on Indian stock market data. Therefore our research involves implementations for Indian stock market. This makes our present study a new case study in the field of forecasting in the Indian stock market. However, this does not limit our resulting analysis and conclusion to our datasets only; instead can be applied to other generic datasets as well. 

To analyse the relative performances of Deep Neural Networks in Time Series Forecasting, we employ the following neural network models:
\begin{enumerate}
    \item Multilayered Network: Multilayer Perceptron (MLP)
    \item Spatial Networks: Convolutional Neural Networks (CNN)
    \item Temporal Networks: Recurrent Neural Networks using: (a) Gated Recurrent Unit (GRU) cells; and (b) Long Short-Term Memory (LSTM) cells
\end{enumerate}
These deep networks are evaluated for two different ways of time series forecasting viz. single-step ahead stock price prediction and multiple-step ahead\footnote{a window of stock prices} stock price prediction. By employing four different state-of-the-art deep network models and with ten different datasets with stock price data from last 17 years, our present work serves as a good case study on the applicability of deep neural networks on Indian stock market data.

\subsection{ Organisation of this paper}
This paper is organised as follows: Section \ref{sec:intro} introduced the motivation, problem statement, and major contributions of this study. In section \ref{sec:relworks}, we provide brief details about research efforts made by the community in the field of statistics and machine learning for time-series forecasting. Section \ref{sec:matmet} provides a detailed description of the data and methodology used by our work. Section \ref{sec:resdis} describes the simulation setup, summarises the results, and discusses the findings. The paper is concluded in section \ref{sec:concl}. The detailed results and time-series prediction plots for various stocks for both one-step as well as multi-step forecasting are provided in Appendix~\ref{appendix:forecasting}.

\section{Related Works}
\label{sec:relworks}

A useful review of multi-step ahead forecasting is published in~\cite{bone2002multi}. These methods describe the different usages of neural networks. They conducted experiments which proposed two constructive algorithms initially developed to learn long-range dependencies in time-series, perform a selective addition of time-delayed to recurrent networks producing noticeable results on single-step forecasting. These results, together with the fact that longer-range delays embodied in the time-delays should be allowed for the system to better learn the series and when predicting for multiple steps and improved results on multi-step prediction problems as can be seen from the experimental evidence. Statistical models are another class of tools suitable and successful for time-series forecasting. One such model is the Autoregressive integrated moving average (ARIMA)~\cite{contreras2003arima}. These models have been quite successful for one-step and sometimes multi-step forecasting.
Further, researchers have explored the idea of hybridising ARIMA and other non-statistical models for forecasting:~\cite{zhang2003time,conejo2005day}. Most successful hybrids are the techniques combining neural networks and statistical models such as as~\cite{zhang2003time,khashei2008new,khashei2011novel}. However, communities continue to explore the comparative domain of statistical model versus neural network models. One of the latest studies on a similar line is work done by Namini and Namini~\cite{siami2018forecasting}, where the authors explore the applicability of ARIMA and LSTM based RNNs. The authors' empirical study on this suggested that deep learning-based algorithms such as LSTM outperform traditional algorithms such as the ARIMA model. More specifically, the average reduction in error rates obtained by LSTM is around 85\% when compared to ARIMA, indicating the superiority of LSTM to ARIMA.

Majumder and Hussian ~\cite{majumder2007forecasting} have used an artificial neural network model with back-propagation to build the network for forecasting. They have studied the effects of hyperparameters, including activation functions. They have critically selected the input variables and have introduced lags between them. They have tried building models with various delays ranging from 1 to 5 day-lags. The input variables chosen for this model are the lagged observation of the closing prices of the NIFTY Index. The experimental results showed that $tanh$ activation function performed better. However, the various day-lags being compared produced varied results based on the loss function used.

Neeraj et al.~\cite{dutta2006artificial} have used Artificial Neural Network (Feedforward Backpropagation Networks) model for modelling BSE Sensex data. After performing initial experiments, a model was finalised, which had 800 neurons with tan-sigmoid transfer function in the input layer, three hidden layers with 600 neurons each, and the output layer with one neuron predicting the stock price. They built two networks. The first used 10-week oscillator and the second one had 5-week volatility. A 10-week oscillator (momentum) is an indicator that gives information regarding the future direction of stock values. When combined with the moving averages, it is observed that it improves the performance of ANN. They used RMSE(Root Mean Squared Error) to calculate errors. They concluded that the first network performed better than the second one for predicting the weekly closing values of BSE Sensex. In a recent study~\cite{hiransha2018nse}, the authors have used different DL architectures like RNNs, LSTMs, CNNs, and MLPs to generate the network for the first dataset where they used TATAMOTORS stock prices for training and have used the trained model to test on stock prices of Maruti, Axis Bank, and HCL Tech. They also built linear models like ARIMA to compare the nonlinear DNN architectures. They made the network having 200 input neurons and ten output neurons. They chose window size as 200 after performing error calculations on various window sizes. They also used this model to test on the other two stocks, which were Bank of America (BAC) and Chesapeake Energy (CHK), to identify the typical dynamics between different stock exchanges. It could be seen from their experimental results that the models were capable of detecting the patterns existing in both the stock markets. Linear models like ARIMA were not able to identify the underlying dynamics within various time series. They concluded that deep architectures (particularly CNNs) performed better than the other networks in capturing the abrupt changes in the system.

Our study is a comprehensive addition to the literature in the sense that this work employs four different deep models for ten different Indian time series data with varying degrees of volatility and significant structural breaks over 17 years. Further, it also explores the performances of such models with regard to one-step and multi-step forecasting. This work could be considered as a significant benchmarking study concerning the Indian stock market.

\section{Materials and Methods}
\label{sec:matmet}

\subsection{Data}
\label{sec:data}
In order to provide generalised inferences and value judgements on the performance of neural networks towards single-step and multi-step time-series forecasting, stock price datasets are quite lucrative as their time-series data typically exhibit characteristics like non-stationarity, multiple structural breaks, as well as high volatility. Further, instead of using a single stock, we used a diversified dataset of 10 different stocks in the Indian stock market. Table~\ref{tab:stocks} describes all the ten stocks that were used for the study. It should be noted that the duration or time-frame of the data for each stock is the same. Furthermore, we use the same dataset of 10 stock prices for both single-step and multi-step forecasting in order to provide better contrasts into the performance of various deep neural network models across both the types of prediction.

\begin{table*}[h]
    \centering
    \caption{Indian stock price data: 10 companies. The period is fixed for all the stocks: 1st January 2002 to 15th Jan 2019 (over 17 years)}
    \label{tab:stocks}
    \begin{tabular}{ll}
        \hline
        Dataset & Description \\ \hline
        ACC & American Campus Communities, Inc. \\
        AXISBANK & Axis Bank Ltd \\
        BHARTIARTL & Bharti Airtel Limited \\
        CIPLA & Cipla Ltd \\
        HCLTECH & HCL Technologies Ltd \\
        HDFC & HDFC Bank Limited \\ 
        INFY & Infosys Ltd. \\
        JSWSTEEL & JSW Steel Limited Fully Paid Ord. Shrs \\
        MARUTI & Maruti Suzuki India Ltd \\
        ULTACEMCO & UltraTech Cement Ltd \\
        \hline
    \end{tabular}
\end{table*}

\subsection{Deep Neural Networks (DNN)}
\label{sec:meths}

We formulate the problem in the following way. Let $\mathbf{x}$  be a time-series defined as $\mathbf{x}$ = $(x_1, \ldots, x_w,\ldots,x_{w+p})$, where $x_i$ represents the stock price at time-step $i$, $w$ refers to window-size and $w_{test}$ refers to the test period for which forecast is to be evaluated. So, a time-steps $(w+1,\ldots,w+w_{test}$ means a $w_{test}$-period window. Correspondingly, we will denote neural network predictions for this $w+1$ to $w+w_{test}$ time-steps as $(\hat{x}_{w+1},\hat{x}_{w+2},\ldots,\hat{x}_{w+w_{test}})$.

For single-step forecasting, the goal is to predict $\hat{x}_{w+1}$ given $(x_1, x_2, \ldots, x_w)$. Mathematically, we can express this as:
\begin{equation}
\hat{x}_{w+1} = f\left((x_1, x_2, \ldots, x_w);\bm{\theta}\right)
\end{equation}
where, $\bm{\theta}$ is the learnable model parameters and $f$ represents a deep network.

Multi-step prediction can be done using two approaches: iterative approach, and direct approach~\cite{hamzaccebi2009comparison}. In iterative method, first subsequent period information is predicted through past observations. Afterwards, the estimated value is used as an input\footnote{the prediction goes as an input feature}; thereby the next period is predicted. The process is carried on until the end of the forecast horizon\footnote{a synonym for forecast window commonly used in time-series community}. The function produces single value at every future time-step. Let $(x_1,\ldots,x_w)$ be the last window of the input time-series, and $(x_{w+1},\ldots,x_{w+w_{test}})$ is the stock values for the forecast horizon $w_{test}$. The goal is to predict $(\hat{x}_{w+1},\ldots,\hat{x}_{w+w_{test}})$. Using iterative approach, this can be defined as follows:
Consider an iterator variable $j \in \{w+1,\ldots,w+w_{test}\}$. If $w+1 \leq j \leq 2w$,
\begin{equation}
    \hat{x}_{j} = f((x_{j-w}, x_{j-w+1}, \ldots, x_{w}, \hat{x}_{w+1}, \ldots, \hat{x}_{j-1}); \bm{\theta});
\end{equation}
and, if $j > 2w$,
\begin{equation}
    \hat{x}_{j} = f((\hat{x}_{j-w}, \hat{x}_{j-w+1}, \ldots, \hat{x}_{j-1}); \bm{\theta})
\end{equation}

In the direct multi-step forecast method, successive periods can be predicted all at once. Each prediction is related only to the stock values in the input window. We can write this as:
\begin{equation}
\hat{x}_{j} = f\left((x_k,\ldots,x_{w + k - 1});\bm{\theta} \right)
\end{equation}
where, $j \in \{w + k,\ldots,w + k - 1 + w_{test}\}$ and
$k$ is a variable used to denote the iterator over
the day instance.

In the following subsections, we briefly describe the existing deep network tools used in this work. These tools are standard, and the mathematical details could be found in the corresponding references, and therefore, we do not explicitly provide the precise mathematical workings of these models.

\subsubsection{Multilayer Perceptron (MLP)}
An MLP consists of at least three layers of nodes: an input layer, a hidden layer, and an output layer~\cite{hastie2005elements}. Except for the input nodes, each node is a neuron that uses a nonlinear activation function. MLP utilises a supervised learning technique called back-propagation for training~\cite{rumelhart1985learning}. The inputs in our case will be time-series data from a specific window.

\subsubsection{Convolutional Neural Network (CNN)}
The idea behind CNNs~\cite{lecun1995convolutional} is to convolve a kernel (whose size can be varied) across an array of input values (like in time series data) and extract features at every step. The kernel convolves along the array based on the stride parameter provided. The stride parameter determines the amount with which the kernel moves along the input to learn the required features for predicting the final output. In our case, we have done 1D convolution on our array of stock prices from various time steps with appropriate kernel size. This kernel learns the features from that window of the input in order to predict the next value as accurately as possible. This technique, however, does not capture time-series co-relations and treats each window size separately.

\subsubsection{Recurrent Neural Network (RNN)}
RNNs make use of sequential information to learn and understand the input features. These are different from MLPs, where inputs and outputs are assumed to be independent. But the conventional methods fail in situations where inputs and outputs influence each other (time-dependence)~\cite{graves2008novel}. RNNs are recurrent as they process all the steps in a sequence in the same way and produce outputs that depend on previous outputs. In other words, RNNs have a memory that stores all the information gained so far. Theoretically, they are expected to learn and remember information from long sequences, but practically, they have found to be storing information only from a few steps back. In our work, we have passed the input time series data sequentially one by one into the network. The hidden states are trained accordingly and are used to predict the next stock price. During training, we compare the predicted and true values and try to reduce the error difference. During testing, we use the previous predicted value to calculate the next time steps (future stock prices). 

\noindent
(a) \textit{Gated-Recurrent Units (GRU) based RNN:}
The principle of both GRU and LSTM~\cite{hochreiter1997long} cells are similar, in the sense that they both are used as "memory" cells and are used to overcome the vanishing gradient problem of RNNs. A GRU cell, however, has a different gating mechanism in which it has two gates, a reset gate, and an update gate~\cite{cho-etal-2014-learning}. The idea behind the reset gate is that it determines how much of the previously gained memory or hidden state needs to be forgotten. The update gate is responsible for deciding how much of the past gained information needs to be passed along the network. The advantage of using the gating mechanism in these cells is to learn long-term dependencies.

\noindent
(b) \textit{Long-Short Term Memory Cells (LSTM) based RNN:}
LSTMs~\cite{hochreiter1997long} cells were designed to overcome the problem of vanishing gradients in RNNs. Vanishing gradients is a problem faced in deeper networks when the error propagated through the system becomes smaller due to which training and updating of weights do not happen efficiently. LSTMs overcome this problem by embedding the gating mechanism in each of their cells. They have input, forget, and output gates which updates and controls the cell states. The input gate is responsible for the amount of new hidden state computed after the current input you wish to pass through the ahead network. The forget network decides how much the previous state it has to let through. In the end, the output gate defines how much of the current state it has to expose to the higher layers (next time steps).

\subsection{Implementation}
For single-step forecasting, the input window (i.e. backcast window) size is studied in the set \{3, 5, 7, 9, 11, 13, 15\}. The implementation for this is straightforward, as explained in section~\ref{sec:meths}, where the testing window is a single stock value in the future. For the multi-step forecasting, the implementation is conducted for 3 different backcast windows \{30, 60, 90\} and
4 different forecast windows such as \{7, 14, 21, 28\}. The implementation for the multi-step forecasting is carried out using the direct strategy as described earlier.

Further, the following details are relevant in our implementations: The original data for prices of all the stocks were normalised to the interval range $[0,1]$. For each stock, the goal was to use the training set for model building, post which the trained model would be used to predict the whole test set. The train-test split for each stock was done in such a way that the training set comprised of stock prices from 1st January 2002 to 1st January 2017, and the subsequent prices formed the testing set.

It should be noted that for all the deep network models, the input size remains equal to the window size ($w$). The deep networks involve 
many different hyperparameters; however, given the amount of data and
computational resources available to us, we were limited to perform some manual tuning of these parameters. Due to reason of space, we are unable to provide these details. We note that automatically tuning various hyperparameters of these deep networks could result in better forecast performance. The manually fixed set of hyperparameter details are furnished below:
\begin{description}
    \item[MLP:~] There are 2 hidden layers with sizes $(16, 16)$. The output layer has 1 neuron. The activation functions in all layers are $\mathtt{relu}$ (rectified linear unit).
    \item[CNN:~] There are 4 hidden layers with sizes $(32,32,2,32)$ with the third layer being a Max-Pooling Layer. The output layer has the size 1. The activation function used in every layer is $\mathtt{relu}$.
    \item[GRU-RNN:~] There are 2 hidden layers with sizes $(256, 128)$. The output layer has 1 neuron. The activation function used in each layer is $\mathtt{relu}$ with $\mathtt{linear}$ activation for
    the final layer.
    \item[LSTM-RNN:~] There are 2 hidden layers with sizes $(256, 128)$. The output layer has 1 neuron. The activation function used for every layer is $\mathtt{relu}$ with $\mathtt{linear}$ activation for the final layer.
\end{description}

The evaluation or loss metric for these models is `mean-squared-error (MSE)'. Further, for reliable model evaluation, and each model was independently run (trained and tested) for 5 different times to obtain statistically reliable performance estimates. Consequently, we obtained results in the form of loss intervals corresponding to our predictions on the test datasets vs the actual stock prices. These testing loss intervals have been reported in the results' tables. These test loss intervals provide a summary in the form of the mean and standard deviation of MSE obtained over five different runs. In the tables, the representation of the loss intervals is $mean~(\pm std. dev.)$.

All our implementations are carried out in the Python environment. The deep neural networks are implemented using the Python library: $\mathsf{Keras}$. All the experiments
are conducted in a machine with Intel i7 processor, 16GB main memory 
and 
NVIDIA 1050 GPU that has 4GB of video memory. We used the Python $\mathsf{nsepy}$ library to fetch the historical data for all Indian stocks from the National Stock Exchange (NSE: \url{https://www.nseindia.com/}). The code and data are shared via
GitHub repository: \url{https://github.com/kaushik-rohit/timeseries-prediction}.

\section{Result and Discussion} \label{sec:resdis}

In this section, we provide a summary of results that are obtained for single- and multi-step forecasting of the 10 different Indian stock data. For clear presentation, we place all the result tables, and some
sample forecast plots in Appendix~\ref{appendix:forecasting} and
only provide the statistical test results in this section. However, the individual forecast result tables are referred to in the discussion text.

\subsection{Single-Step Forecasting} 

The performance observed for the ACC stock depicts that all four deep models seem to perform the prediction task similarly. However, as we increase the window sizes, the predictions of all the models go further away from the true values increasing the error rate. Hence, it can be concluded that the future single stock price is highly dependent on the immediate previous prices and less dependent on further past prices. However, a different kind of prediction trend was shown by the models for the AXISBANK stocks. It can be seen from the graphs of AXISBANK stocks that all the models performed quite well for the smallest window-size of 3. The predictions for the window-size 7 were also good for all the models. However, the results for the other window-sizes varied irregularly and didn't perform as well. A very different trend was seen for BHARTIARTL stock prediction. Table~\ref{tab:bharti_sf} suggests that for smaller window-sizes, MLP performed slightly better than the others. However, as window-sizes increases, CNN starts outperforming all the other models. One unique aspect of these models can be observed in the forecasting graphs (refer Appendix~\ref{appendix:forecasting}): all the models failed to predict the sudden increases in the prices to the actual extent. Hence, it could be emphasised that the information from the previous trends of stock prices is not sufficient enough for predicting future prices, and thus, it may depend on a variety of factors that have not been incorporated in these models. A filter-based deep network such as CNN outperforms other deep models for CIPLA stock dataset as shown in Table~\ref{tab:cipla_sf}. This holds for all window sizes.
However, the results obtained for the HCLTECH stock is quite contradictory. Table~\ref{tab:hcl_sf} represents that GRU-RNN performs much better compared to other models. The window-size of 13 produced the best result within the GRU model. This demonstrates that the GRU-RNN structure could certainly handle the deviation within the stock prices for an extended period (i.e., $w=13$). The almost similar inference could also be made for HDFC stock, where both LSTM-RNN and GRU-RNN have performed very well for $w=9$ (refer Table~\ref{tab:hdfc_sf}). Table~\ref{tab:infy_sf} shows that an identical trend in performance was observed across different window sizes for INFY stock price prediction. Additionally, CNN required a higher number of input features (i.e., $w=11$) to perform to its capacity for this dataset. 

The JSWSTEEL stock dataset contains a very high number of structural breaks and is highly volatile. Table~\ref{tab:jsw_sf} shows that this characteristic behaved as an adversarial feature for all the models, and hence the models were not able to perform well. However, LSTM-RNN shows some improved performance given a higher input window of 13. Table~\ref{tab:maruti_sf} suggests that a similar trend in performance was also observed for the MARUTI stock dataset with a surprising result that the model like MLP could perform better than other deep models with minimal input window of 3. MLP also performs better than its counterparts for the ULTRACEMCO dataset, as shown in Table~\ref{tab:ultra_sf}.

\subsubsection{Statistical significance test} 

The results obtained over five different independent runs of the models are subjected to a statistical significance test. For this, we conduct the
Diebold-Mariano test~\cite{diebold1995comparing,harvey1997testing}. However,
we conduct the DM-test only for the single-step forecasting results. The DM-test
compares two hypotheses at a time, and the value is converted into the $p$-value.
From Table~\ref{tab:dmtest}, it could be concluded that most of the results are
significant given any hypotheses pair. The results of Diebold-Mariano Test at 0.01\% level of significance ($\alpha= 0.0001$) suggests that the relative order
of performance of the deep network models for single-step
forecasting is: GRU-RNN, CNN, LSTM-RNN and MLP, where
MLP outperforms all others. We note that
the statistical significance strongly looks at overall
performance of the model rather than the performance on
individual dataset. Althoguh, MLP does not encode
any long-term dependency arising in the time-series
data, it may not be expected to perform as good
as standard dependency-learning models such as
LSTM- or GRU-RNNs. Another observation that could be made is that the data used in our present work may not be
containing any such long-term dependencies for which
a sequence-based deep model or a convolution-based deep model could be very useful. Our goal here is not to 
recommend MLP as the best model for real-world 
applications to time-series modelling, rather as
a typical deep model that performs well on
data that has mutiple structural breaks and is highly
volatile. However, readers should note that the level of
significance plays a crucial role in choosing the performance ordering of the models.
\begin{table*}[!h]
    \caption{Statistical significance test for single step forecasting results. The table shows the value of DM-statistic followed by the corresponding p-value within parenthesis.}
    \label{tab:dmtest}
    \centering
    \footnotesize{
    \begin{tabular}{cccccc}
        \hline
        Stocks & MLP-CNN & LSTM-GRU & MLP-LSTM & LSTM-CNN & CNN-GRU\\ \hline
        ACC        & -1.6825 (0.09) & -2.2266 (0.02) & 1.9484 (0.05) & 1.1225 (0.26) & -1.4563 (0.14)\\ \hline
        AXISBANK   & 1.0484 (0.29) & 4.3022 (0.02e-3) & -4.2013 (0.03e-3) & 4.2392 (0.02e-3) & 1.0799 (0.28)\\ \hline
        BHARTIARTL & 3.2570 (0.00) & -3.5133 (0.00) & -2.3957 (0.01) & 2.7976 (0.00) & -3.4586 (0.00)\\ \hline
        CIPLA      & -0.0928 (0.92) & -3.5751 (0.00) & -3.9925 (0.07e-3) & 3.8955 (0.00) & -4.3625 (0.15e-4)\\ \hline
        HCLTECH    & -5.6934 (0.02e-6) & 6.2360 (0.09e-8) & -6.2722 (0.07e-8) & 6.4186 (0.03e-8) & 5.5730 (0.41e-8)\\ \hline
        HDFC       & -1.8418 (0.06) & 1.1548 (0.24) & -0.1273 (0.89) & -0.5156 (0.60) & 2.2790 (0.02)\\ \hline
        INFY       & -0.8889 (0.37) & 1.3987 (0.16) & -1.1016 (0.27) & 1.2804 (0.20) & -1.0152 (0.31)\\ \hline
        JSWSTEEL   & 0.9842 (0.32) & 1.0997 (0.27) & 0.9799 (0.32) & -0.9794 (0.32) & 1.0017 (0.31) \\ \hline
        MARUTI     & 0.6225 (0.53) & -7.6933 (0.07e-12) & -2.2788 (0.02) & 2.2733 (0.02) & -7.6887 (0.80e-13)\\ \hline
        ULTRACEMCO & -1.3570 (0.17) & -1.5115 (0.13) & -0.2501 (0.80) & -1.5288 (0.12) & -1.4971 (0.13)\\ \hline
    \end{tabular}}
\end{table*}

\subsection{Multi-step Forecasting} 

Multi-step forecasting has always been a challenging problem in time-series prediction problems. The results are in Table~\ref{appendix:mf}.
For Table~\ref{tab:acc_mf} through to Table~\ref{tab:ultra_mf}, the multi-step forecast results suggest that for small forecast window
the deep network methods are performing well for all the datasets. As the forecast window size is increased (such as 28), the performance drops significantly. The performance of the four deep network models
for the ACC stock data suggests that the MLP needs to observe
as high as 30 input days to predict accurately 7 days
of future data. This is expected for a densely connected
network like an MLP where the salient features are constructed in its intermediate hidden layers. This observation also holds for other stocks expect for the JSWSTEEL stocks.
Furthermore, it is in contradiction to more inputs as 60 or 90, where additional days don't aid any useful information to the model. Similarly, for the JSWSTEEL stocks, the performance for the MLP model is best at
60 input days to produce 7 days ahead forecast of stock
prices.

The GRU-RNN model looks into a large sized input such as
60 or 90 to make predictions for 7 days in the future, whereas for the LSTM-RNN and CNN, 30 days of input is sufficient to produce accurate future predictions. Similarly, 
looking at all the performance models for all possible
forecast windows considered in this work such as \{7, 14, 21, 28\}, we note that MLP outperforms all other deep models for the majority of stocks. To support the observation, we conduct a statistical significance test for a sample input-output combination. 

\subsubsection{Statistical significance test}

The DM test results for multi step forecasting with input window size 30 and output window size 7 is in Table ~\ref{tab:dmtest2}. The level of significance is set at 0.1\%. For comparing the relative forecasting performance of any pair of models from the table, we take a majority vote based on DM-test analysis for each of the 10 stocks. Accordingly, for each pair of model comparison, one model is chosen as the best among the pair if it is found to be the best model for more than 5 out of 10 stocks based on the DM-test p-value analysis for that pair of models. It is observed that MLP outperforms all the other deep network approaches for this setting of input and output window combination. This observation is consistent with the observation for the single-step forecasting performance as well. The overall order of relative forecasting performance of different neural networks for multi-step forecasting is found to be: CNN, LSTM-RNN, GRU-RNN, and MLP. Readers should note that the level of
significance plays a crucial role in choosing the performance ordering of the models.
\begin{table*}[!h]
    \caption{Statistical significance test for multi step forecasting results with input window size 30 and output window size 7. The table shows the value of DM-statistic followed by the corresponding p-value within parenthesis.}
    \label{tab:dmtest2}
    \centering
    \footnotesize{
    \begin{tabular}{cccccc}
        \hline
        Stocks & MLP-CNN & LSTM-GRU & MLP-LSTM & LSTM-CNN & MLP-GRU\\ \hline
        ACC        & -2.7965 (0.01) & -2.4170 (0.02) & 1.9168 (0.06) & -2.6386 (0.01) & -1.3845 (0.16)\\ \hline
        AXISBANK   & -1.6748 (0.09) & -2.9254 (0.03e-1) & -2.8932 (0.03e-1) & 2.5876 (0.01e-1) & -2.9673 (0.00)\\ \hline
        BHARTIARTL & -2.2470 (0.03) & 3.5146 (0.00) & -2.3641 (0.02) & -0.4689 (0.64) & 0.4332 (0.66)\\ \hline
        CIPLA      & -1.9501 (0.05) & 3.3701 (0.01e-2) & 0.1213 (0.90) & -1.8521 (0.06) & 0.7197 (0.47)\\ \hline
        HCLTECH    & -6.5086 (1.95e-10) & 7.8612 (0.02e-10) & -7.8557 (0.02e-12) & -5.3166 (1.63e-14) & -6.6640 (0.75e-10)\\ \hline
        HDFC       & -2.4685 (0.01) & 4.4679 (0.09e-2) & -0.0014 (0.99) & -2.2448 (0.02) & 2.8395 (0.00)\\ \hline
        INFY       & 2.5914 (0.01) & 2.0002 (0.05) & -2.3651 (0.02) & 2.4460 (0.01) & -2.3381 (0.01)\\ \hline
        JSWSTEEL   & 0.8546 (0.39) & 8.4322 (4.23e-16) & 1.3509 (0.18) & -2.7461 (0.01) & 1.6018 (0.10)\\ \hline
        MARUTI     & -1.8351 (0.07) & -10.1729 (4.27e-22) & -5.6069 (3.52e-8) & 5.5603 (4.53e-8) & -10.1630 (0.46e-21)\\ \hline
        ULTRACEMCO & -1.7387 (0.08) & -1.5022 (0.13) & 1.5481 (0.12) & -2.0443 (0.04) & 0.3046 (0.76)\\ \hline
    \end{tabular}}
\end{table*}

\section{Conclusion} \label{sec:concl}

In this paper, we studied the applicability of the popular deep neural networks (DNN) comprehensively as function approximators for non-stationary time-series forecasting. Specifically, we evaluated the following DNN models: Multi-layer Perceptron (MLP), Convolutional Neural Network (CNN), RNN with Long-Short Term Memory Cells (LSTM-RNN), and RNN with Gated-Recurrent Unit (GRU-RNN). These four powerful DNN methods have been evaluated over ten popular Indian financial stocks' datasets. Further, the evaluation is carried out through predictions in both fashions: (1) single-step-ahead, (2) multi-step-ahead. The training of the deep models for both single-step and multi-step forecasting has been carried out using over 15 years of data and tested on two years of data. Our experiments show the following: (1) The neural network models used in this experiments demonstrate good predictive performance for the case of single-step forecasting across all stocks datasets; (2) the predictive performance of these models remains consistent across various forecast window sizes; and (3) given the limited input window condition for multi-step forecasting, the performance of the deep network models are not as good as that was seen in the case of single-step forecasting. However, notwithstanding the above limitation of the models for the multi-step forecasting, given the vast amount of data collected over a duration of 17 years on which the models are built, this work could be considered as a significant benchmark study with regard to the Indian stock market. Further, we note the following observation. The deep network models are built with raw
time-series of stock prices. That is: no external features such as micro- or macro-economic factors, other statistically handcrafted parameters, relevant news data are provided to these models. These parameters are often considered to be useful to impact stock price prediction. A model that takes into account these additional factors could better the predictive performance of both single-step as well as multi-step forecasting.

\bibliographystyle{unsrt}
\bibliography{references}

\pagebreak

\appendix

\section{Appendix: Forecasting Results}
\label{appendix:forecasting}

The forecasting plots during testing. The plots are showing the average of five independent runs of the programs, and this average is compared with the true value. Due to the reason of space, we provide results for only one stock dataset (ACC). However, similar performances were also observed for other datasets, which can be located in the link \url{https://github.com/kaushik-rohit/timeseries-prediction}.

\setcounter{figure}{0}
\setcounter{table}{0}
\renewcommand{\thetable}{A\arabic{table}}
\renewcommand{\thefigure}{A\arabic{figure}}
\label{appendix:sf}
\setcounter{figure}{0}    
\begin{table*}[h]
    \centering
    \caption{Single-step forecast results for ACC stock: Test MSE (mean and std. dev. over 5 different runs)}
    \label{tab:acc_sf}
    \footnotesize
    \begin{tabular}{ccccc}
        \hline
        Window-sizes & MLP & CNN & GRU-RNN & LSTM-RNN \\ \hline
        3 & 0.0004221 (0.0000444) & 0.0005847 (0.0000562) & 0.0016516 (0.0008691) & 0.0015011(0.0008549) \\ \hline
        5 & 0.0006502 (0.0002251) & 0.0007508 (0.0001441) & 0.0016215 (0.0006634) & 0.0016570 (0.0013026) \\ \hline
        7 & 0.0008696 (0.0002147) & 0.0011275 (0.0004138) & 0.0026603 (0.0010596) & 0.0017234 (0.0010127) \\ \hline
        9 & 0.0010079 (0.0001728) & 0.0011650 (0.0002645) & 0.0030460 (0.0009964) & 0.0021387 (0.0014420) \\ \hline
        11 & 0.0010929 (0.0002465) & 0.0011588 (0.0000686) & 0.0027522 (0.0015047) & 0.0021330 (0.0010239) \\ \hline
        13 & 0.0012379 (0.0001855) & 0.0015122 (0.0001267) & 0.0036655 (0.0009325) & 0.0014072 (0.0008664) \\ \hline
        15 & 0.0017524 (0.0003690) & 0.0017017 (0.0003527) & 0.0039116 (0.0023042) & 0.0020953 (0.0005682) \\ \hline
    \end{tabular}
\end{table*}

\begin{table*}[h]
    \centering
    \caption{Single-step forecast results for AXISBANK stock: Test MSE (mean and std. dev. over 5 different runs)}
    \label{tab:axis_sf}
    \footnotesize
    \begin{tabular}{ccccc}
        \hline
        Window-sizes & MLP & CNN & GRU-RNN & LSTM-RNN \\ \hline
        3 & 0.0000515(0.0000260) & 0.0000487(0.0000217) & 0.0004417(0.0004302) & 0.0003082(0.0003509) \\ \hline
        5 & 0.0000693(0.0000336) & 0.0000771(0.0000287) & 0.0001848(0.0001252) & 0.0005975(0.0002366) \\ \hline
        7 & 0.0000690 (0.0000313) & 0.0001497(0.0000982) & 0.0003395(0.0001962) & 0.0001344(0.0000540) \\ \hline
        9 & 0.0000811(0.0000116) & 0.0001389(0.0000653) & 0.0003509 (0.0004081) & 0.0004543(0.0003342) \\ \hline
        11 & 0.0001067(0.0000279) & 0.0000850(0.0000448) & 0.0005067(0.0006814) & 0.0001788 (0.0001291) \\ \hline
        13 & 0.0000782(0.0000347) & 0.0000722(0.0000281) & 0.0003205(0.0002610) & 0.0005240(0.0001061) \\ \hline
        15 & 0.0001111(0.0000170) & 0.0001805 (0.0000904) & 0.0004173(0.0005002) & 0.0002738(0.0002042) \\ \hline
    \end{tabular}
\end{table*}

\begin{table*}[h]
    \centering
    \caption{Single-step forecast results for BHARTIARTL stock: Test MSE (mean and std. dev. over 5 different runs)}
    \label{tab:bharti_sf}
    \footnotesize
    \begin{tabular}{ccccc}
        \hline
        Window-sizes & MLP & CNN & GRU-RNN & LSTM-RNN \\ \hline
        3 & 0.0002035(0.0000681) & 0.0002797 (0.0001659) & 0.0005684 (0.0000984) & 0.0004817(0.0001587) \\ \hline
        5 & 0.0002316(0.0001263) & 0.0002699(0.0001950) & 0.0015586 (0.0002458) & 0.0005880 (0.0001752) \\ \hline
        7 & 0.0003627(0.0000733) & 0.0004076(0.0002003) & 0.0011911 (0.0005273) & 0.0005556 (0.0002353) \\ \hline
        9 & 0.0004252(0.0001691) & 0.0003931(0.0002384) & 0.0014297 (0.0006784) & 0.0007216 (0.0001412) \\ \hline
        11 & 0.0005564 (0.0000862) & 0.0005313 (0.0002783) & 0.0018257 (0.0005343) & 0.0008124 (0.0003259) \\ \hline
        13 & 0.0004261 (0.0002045) & 0.0003093 (0.0001621) & 0.0011227 (0.0004965) & 0.0006211 (0.0001480) \\ \hline
        15 & 0.0007215 (0.0001667) & 0.0002497 (0.0001458) & 0.0010685 (0.0008361) & 0.0007042 (0.0003836) \\ \hline
    \end{tabular}
\end{table*}

\begin{table*}[h]
    \centering
    \caption{Single-step forecast results for CIPLA stock: Test MSE (mean and std. dev. over 5 different runs)}
    \label{tab:cipla_sf}
    \footnotesize
    \begin{tabular}{ccccc}
        \hline
        Window-sizes & MLP & CNN & GRU-RNN & LSTM-RNN \\ \hline
        3 & 0.0000871 (0.0000337) & 0.0001130 (0.0000594) & 0.0009330 (0.0004240) & 0.0001653 (0.0000617) \\ \hline
        5 & 0.0000856 (0.0000145) & 0.0001262 (0.0000551) & 0.0012812 (0.0006000) & 0.0003738 (0.0002768) \\ \hline
        7 & 0.0001197 (0.0000311) & 0.0001435 (0.0000379) & 0.0009877 (0.0008997) & 0.0004203 (0.0002756) \\ \hline
        9 & 0.0001462 (0.0000355) & 0.0001614 (0.0000390) & 0.0010927 (0.0003391) & 0.0008656 (0.0007267) \\ \hline
        11 & 0.0001899 (0.0000823) & 0.0001907 (0.0000721) & 0.0007634 (0.0002993) & 0.0003982 (0.0001604) \\ \hline
        13 & 0.0001976 (0.0001014) & 0.0002209 (0.0000843) & 0.0014596 (0.0012414) & 0.0010129 (0.0004355) \\ \hline
        15 & 0.0002379 (0.0000800) & 0.0001144 (0.0000328) & 0.0013618 (0.0005843) & 0.0007791(0.0001930) \\ \hline
    \end{tabular}
\end{table*}

\begin{table*}[h]
    \centering
    \caption{Single-step forecast results for HCLTECH stock: Test MSE (mean and std. dev. over 5 different runs)}
    \label{tab:hcl_sf}
    \footnotesize
    \begin{tabular}{ccccc}
        \hline
        Window-sizes & MLP & CNN & GRU-RNN & LSTM-RNN \\ \hline
        3 & 0.0016502 (0.0011387) & 0.0023050 (0.0023157) & 0.0004055 (0.0003749) & 0.0009085 (0.0004147) \\ \hline
        5 & 0.0029187 (0.0026613) & 0.0034353 (0.0027634) & 0.0003305 (0.0002079) & 0.0026182 (0.0026467) \\ \hline
        7 & 0.0019523 (0.0024012) & 0.0022142 (0.0016357) & 0.0015353 (0.0014942) & 0.0087284 (0.0032211) \\ \hline
        9 & 0.0032018 (0.0019762) & 0.0057737 (0.0032285) & 0.0006866 (0.0006232) & 0.0076727 (0.0056202) \\ \hline
        11 & 0.0051562 (0.0020755) & 0.0053916 (0.0030622) & 0.0003930 (0.0002421) & 0.0042450 (0.0025408) \\ \hline
        13 & 0.0022825 (0.0018382) & 0.0071111 (0.0020174) & 0.0001733 (0.0000372) & 0.0075984 (0.0015168) \\ \hline
        15 & 0.0050976 (0.0028671) & 0.0081210 (0.0038284) & 0.0018683 (0.0023934) & 0.0048502 (0.0042392) \\ \hline
    \end{tabular}
\end{table*}

\begin{table*}[h]
    \centering
    \caption{Single-step forecast results for HDFC stock: Test MSE (mean and std. dev. over 5 different runs)}
    \label{tab:hdfc_sf}
    \footnotesize
    \begin{tabular}{ccccc}
        \hline
        Window-sizes & MLP & CNN & GRU-RNN & LSTM-RNN \\ \hline
        3 & 0.0001835 (0.0000824) & 0.0007788 (0.0005064) & 0.0016444 (0.0021480) & 0.0005317 (0.0003852) \\ \hline
        5 & 0.0003086 (0.0003135) & 0.0005371 (0.0001659) & 0.0013700 (0.0015615) & 0.0010208 (0.0010850) \\ \hline
        7 & 0.0006299 (0.0005823) & 0.0005523 (0.0002540) & 0.0012404 (0.0008231) & 0.0006913 (0.0006490) \\ \hline
        9 & 0.0006540 (0.0002297) & 0.0008980 (0.0000963) & 0.0014943 (0.0016051) & 0.0005813 (0.0008020) \\ \hline
        11 & 0.0008674 (0.0003254) & 0.0008349 (0.0002675) & 0.0003954 (0.0002056) & 0.0010702 (0.0014710) \\ \hline
        13 & 0.0007940 (0.0003468) & 0.0006309 (0.0003891) & 0.0012609 (0.0013088) & 0.0023305 (0.0012481) \\ \hline
        15 & 0.0007099 (0.0001384) & 0.0010326 (0.0008803) & 0.0026690 (0.0025715) & 0.0020410 (0.0017725) \\ \hline
    \end{tabular}
\end{table*}

\begin{table*}[h]
    \centering
    \caption{Single-step forecast results for INFY stock: Test MSE (mean and std. dev. over 5 different runs)}
    \label{tab:infy_sf}
    \footnotesize
    \begin{tabular}{ccccc}
        \hline
        Window-sizes & MLP & CNN & GRU-RNN & LSTM-RNN \\ \hline
        3 & 0.0005255 (0.0001396) & 0.0002269 (0.0000797) & 0.0001357 (0.0000900) & 0.0004860 (0.0003214) \\ \hline
        5 & 0.0005674 (0.0001610) & 0.0003122 (0.0001724) & 0.0003263 (0.0002350) & 0.0002029 (0.0001462) \\ \hline
        7 & 0.0004635 (0.0001925) & 0.0002579 (0.0001686) & 0.0001498 (0.0000581) & 0.0002305 (0.0000931) \\ \hline
        9 & 0.0006265 (0.0002582) & 0.0002504 (0.0001102) & 0.0000956 (0.0000186) & 0.0001955 (0.0000614) \\ \hline
        11 & 0.0007612 (0.0001806) & 0.0003764 (0.0002317) & 0.0001351 (0.0000309) & 0.0001972 (0.0000733) \\ \hline
        13 & 0.0007907 (0.0001193) & 0.0004738 (0.0001561) & 0.0002171 (0.0000975) & 0.0001739 (0.0000517) \\ \hline
        15 & 0.0007154 (0.0003186) & 0.0005986 (0.0003641) & 0.0001353 (0.0000668) & 0.0002354 (0.0000706) \\ \hline
    \end{tabular}
\end{table*}

\begin{table*}[h]
    \centering
    \caption{Single-step forecast results for JSWSTEEL stock: Test MSE (mean and std. dev. over 5 different runs)}
    \label{tab:jsw_sf}
    \footnotesize
    \begin{tabular}{ccccc}
        \hline
        Window-sizes & MLP & CNN & GRU-RNN & LSTM-RNN \\ \hline
        3 & 0.0008873 (0.0002596) & 0.0030845 (0.0004266) & 0.0022322 (0.0002757) & 0.0021728 (0.0001631) \\ \hline
        5 & 0.0006888 (0.0004107) & 0.0013932 (0.0005308) & 0.0006842 (0.0003586) & 0.0013430 (0.0004050) \\ \hline
        7 & 0.0006232 (0.0003326) & 0.0017431 (0.0009391) & 0.0016630 (0.0010976) & 0.0005379 (0.0001549) \\ \hline
        9 & 0.0008845 (0.0004175) & 0.0013794 (0.0007401) & 0.0009644 (0.0004872) & 0.0003038 (0.0001008) \\ \hline
        11 & 0.0012099 (0.0005770) & 0.0018192 (0.0009440) & 0.0006822 (0.0004396) & 0.0004553 (0.0001950) \\ \hline
        13 & 0.0010368 (0.0004484) & 0.0012310 (0.0006584) & 0.0009355 (0.0003718) & 0.0009499 (0.0010030) \\ \hline
        15 & 0.0014731 (0.0005437) & 0.0015114 (0.0008925) & 0.0008119 (0.0007530) & 0.0010866 (0.0011773) \\ \hline
    \end{tabular}
\end{table*}

\begin{table*}[h]
    \centering
    \caption{Single-step forecast results for MARUTI stock: Test MSE (mean and std. dev. over 5 different runs)}
    \label{tab:maruti_sf}
    \footnotesize
    \begin{tabular}{ccccc}
        \hline
        Window-sizes & MLP & CNN & GRU-RNN & LSTM-RNN \\ \hline
        3 & 0.0022925 (0.0006457) & 0.0006474 (0.0000423) & 0.0366665 (0.0112447) & 0.0136058 (0.0080959) \\ \hline
        5 & 0.0021511 (0.0021414) & 0.0008467 (0.0000926) & 0.0318909 (0.0267657) & 0.0208617 (0.0050294) \\ \hline
        7 & 0.0009793 (0.0001314) & 0.0009672 (0.0000539) & 0.0928953 (0.0106618) & 0.0183959 (0.0164570) \\ \hline
        9 & 0.0020073 (0.0014313) & 0.0010510 (0.0001043) & 0.0802620 (0.0406923) & 0.0349464 (0.0266224) \\ \hline
        11 & 0.0013546 (0.0005229) & 0.0013423 (0.0003510) & 0.0766645 (0.0218722) & 0.0270535 (0.0174924) \\ \hline
        13 & 0.0015955 (0.0006773) & 0.0013866 (0.0002926) & 0.0727358 (0.0558740) & 0.0619624 (0.0391621) \\ \hline
        15 & 0.0019786 (0.0009192) & 0.0022918 (0.0004053) & 0.0743994 (0.0463265) & 0.0304260 (0.0264045) \\ \hline
    \end{tabular}
\end{table*}

\begin{table*}[h]
    \centering
    \caption{Single-step forecast results for ULTRACEMCO stock: Test MSE (mean and std. dev. over 5 different runs)}
    \label{tab:ultra_sf}
    \footnotesize
    \begin{tabular}{ccccc}
        \hline
        Window-sizes & MLP & CNN & GRU-RNN & LSTM-RNN \\ \hline
        3 & 0.0003553 (0.0000743) & 0.0004674 (0.0000524) & 0.0011129 (0.0006957) & 0.0010138 (0.0006587) \\ \hline
        5 & 0.0005149 (0.0001737) & 0.0005909 (0.0000670) & 0.0033692 (0.0024419) & 0.0013757 (0.0012501) \\ \hline
        7 & 0.0006438 (0.0000922) & 0.0007646 (0.0000962) & 0.0018719 (0.0009955) & 0.0008442 (0.0003019) \\ \hline
        9 & 0.0009227 (0.0002358) & 0.0008155 (0.0001067) & 0.0027970 (0.0020777) & 0.0010058 (0.0004020) \\ \hline
        11 & 0.0010645 (0.0001750) & 0.0009565 (0.0001088) & 0.0058653 (0.0053221) & 0.0005961 (0.0002074) \\ \hline
        13 & 0.0010508 (0.0001368) & 0.0010494 (0.0001358) & 0.0060068 (0.0052101) & 0.0018897 (0.0014145) \\ \hline
        15 & 0.0010278 (0.0001846) & 0.0008928 (0.0000935) & 0.0017564 (0.0015364) & 0.0026440 (0.0029301) \\ \hline
    \end{tabular}
\end{table*}

\begin{figure*}[h]
    \subfloat[$w = 3$]{\includegraphics[width = 3in, height = 1.6in]{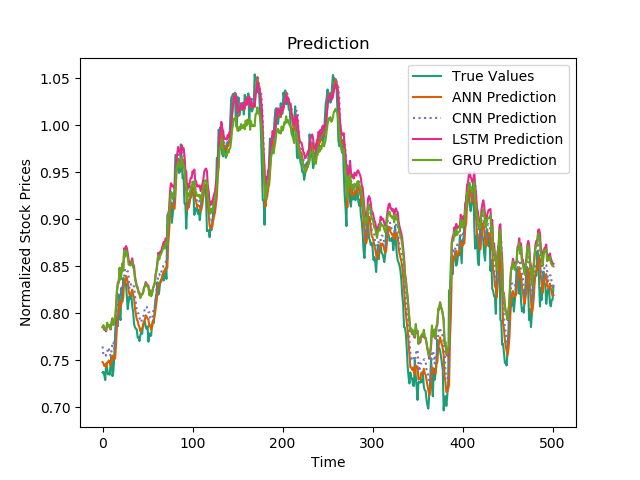}} 
    \subfloat[$w = 7$]{\includegraphics[width = 3in, height = 1.6in]{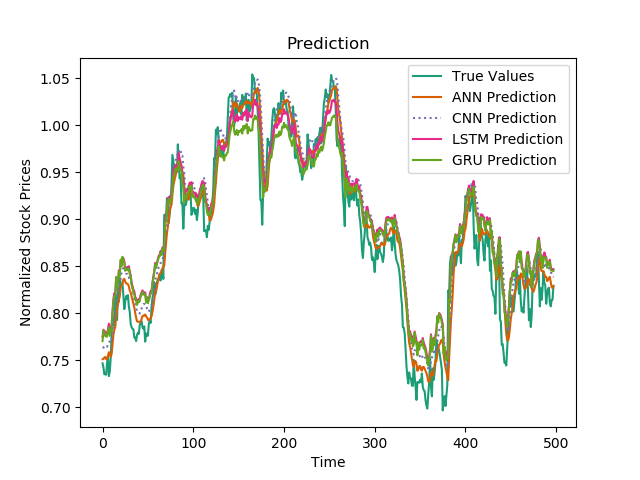}}\\
    \subfloat[$w = 11$]{\includegraphics[width = 3in, height = 1.6in]{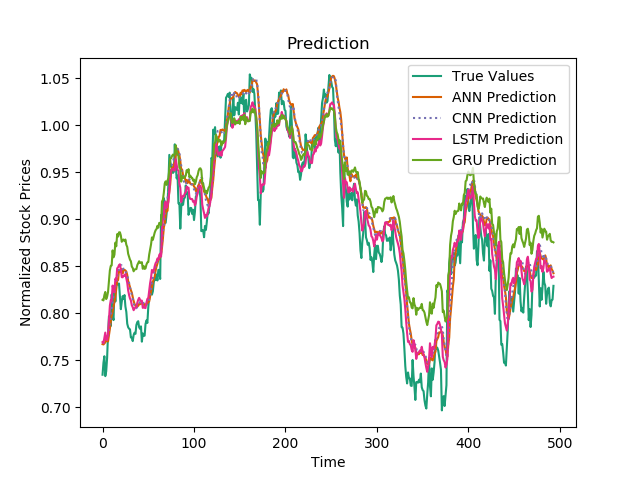}}
    \centering
    \subfloat[$w = 15$]{\includegraphics[width = 3in, height = 1.6in]{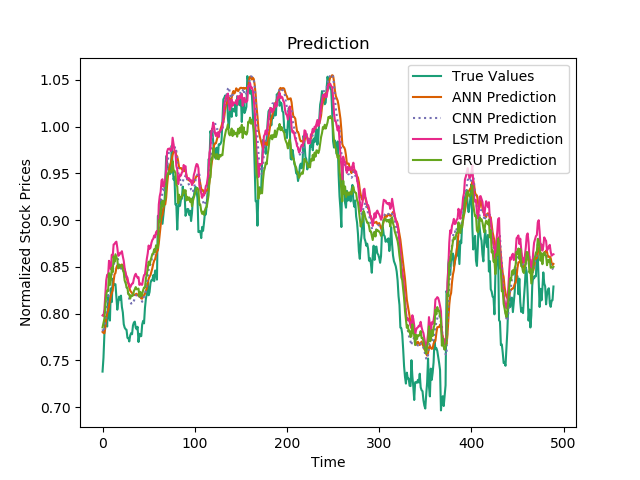}}
    \caption{Single-step forecasting plots for ACC dataset for each deep model against the true prices for different window size ($w$). In the figure legend, ANN refers to MLP. Due to reason of space, we are unable to provide the plots for $w = \{5, 9, 13\}$.}
\end{figure*}

\setcounter{figure}{0}
\setcounter{table}{0}
\renewcommand{\thetable}{B\arabic{table}}
\renewcommand{\thefigure}{B\arabic{figure}}
\label{appendix:mf}
\setcounter{figure}{0}  

\begin{table*}[h]
    \caption{Multi-step forecast results for ACC stock: Test MSE (mean and std. dev. over 5 different runs)}
    \label{tab:acc_mf}
    \centering
    \footnotesize
    \begin{tabular}{ccccc}
        \hline
        (Input, Output) window & MLP & CNN & GRU-RNN & LSTM-RNN \\ \hline
        (30, 7)  & 0.0019233 (0.0001796) & 0.0032655 (0.0007156) & 0.0028288 (0.0008798) & 0.0014831 (0.0001296) \\
        (30, 14) & 0.0023715 (0.0001695) & 0.0034703 (0.0008722) & 0.0024650 (0.0008660) & 0.0024763 (0.0006420) \\
        (30, 21) & 0.0033350 (0.0003381) & 0.0044773 (0.0005973) & 0.0028721 (0.0000534) & 0.0039920 (0.0006566) \\ 
        (30, 28) & 0.0035716 (0.0002659) & 0.0043508 (0.0006285) & 0.0042296 (0.0006304) & 0.0054476 (0.0006860) \\ \hline
        (60, 7)  & 0.0020720 (0.0001856) & 0.0028166 (0.0006529) & 0.0024679 (0.0007471) & 0.0014707 (0.0001257) \\ 
        (60, 14) & 0.0028144 (0.0002397) & 0.0041619 (0.0009518) & 0.0030039 (0.0007646) & 0.0023017 (0.0002294) \\ 
        (60, 21) & 0.0037724 (0.0002339) & 0.0053926 (0.0009353) & 0.0035389 (0.0008466) & 0.0036483 (0.0003987) \\
        (60, 28) & 0.0045590 (0.0003729) & 0.0047837 (0.0005719) & 0.0044799 (0.0000797) & 0.0051585 (0.0005595) \\ \hline
        (90, 7)  & 0.0022322 (0.0002161) & 0.0022918 (0.0002077) & 0.0025558 (0.0008330) & 0.0013354 (0.0001188) \\ 
        (90, 14) & 0.0031306 (0.0004378) & 0.0038712 (0.0006698) & 0.0028143 (0.0003417) & 0.0021925 (0.0002722) \\ 
        (90, 21) & 0.0036370 (0.0006305) & 0.0054046 (0.0015030) & 0.0042265 (0.0014142) & 0.0034945 (0.0006990) \\ 
        (90, 28) & 0.0039776 (0.0003342) & 0.0050536 (0.0007407) & 0.0067699 (0.0012256) & 0.0053037 (0.0012531) \\ \hline
    \end{tabular}
\end{table*}

\begin{table*}[h]
    \caption{Multi-step forecast results for AXISBANK stock: Test MSE (mean and std. dev. over 5 different runs)}
    \label{tab:axis_mf}
    \centering
    \footnotesize
    \begin{tabular}{ccccc}
        \hline
        (Input, Output) window & MLP & CNN & GRU-RNN & LSTM-RNN \\ \hline
        (30, 7)  & 0.0004412 (0.0002132) & 0.0003427 (0.0001689) & 0.0007955 (0.0002145) & 0.0005701 (0.0002627) \\ 
        (30, 14) & 0.0006175 (0.0003172) & 0.0005083 (0.0004398) & 0.0003222 (0.0001544) & 0.0006500 (0.0002760) \\ 
        (30, 21) & 0.0010073 (0.0005057) & 0.0011542 (0.0009714) & 0.0002325 (0.0000245) & 0.0004799 (0.0000963) \\ 
        (30, 28) & 0.0005589 (0.0002897) & 0.0014537 (0.0009028) & 0.0004072 (0.0000335) & 0.0005838 (0.0001089) \\ \hline
        (60, 7)  & 0.0007062 (0.0002869) & 0.0007021 (0.0003361) & 0.0005125 (0.0003105) & 0.0006901 (0.0002465) \\ 
        (60, 14) & 0.0010972 (0.0003644) & 0.0020001 (0.0010282) & 0.0005726 (0.0002575) & 0.0009651 (0.0008234) \\ 
        (60, 21) & 0.0009101 (0.0006839) & 0.0020065 (0.0003630) & 0.0003826 (0.0000839) & 0.0006265 (0.0002844) \\ 
        (60, 28) & 0.0010542 (0.0005515) & 0.0025980 (0.0009502) & 0.0009835 (0.0003565) & 0.0006103 (0.0000680) \\ \hline
        (90, 7)  & 0.0012354 (0.0001973) & 0.0013462 (0.0007233) & 0.0007331 (0.0007168) & 0.0006744 (0.0000642) \\ 
        (90, 14) & 0.0007307 (0.0003338) & 0.0032691 (0.0010425) & 0.0006951 (0.0003462) & 0.0007027 (0.0002224) \\ 
        (90, 21) & 0.0007812 (0.0005256) & 0.0020593 (0.0013153) & 0.0008963 (0.0005536) & 0.0005048 (0.0004352) \\ 
        (90, 28) & 0.0012692 (0.0004672) & 0.0028267 (0.0018528) & 0.0013811 (0.0010722) & 0.0007381 (0.0002818) \\ \hline
    \end{tabular}
\end{table*}

\begin{table*}[h]
    \caption{Multi-step forecast results for BHARTIARTL stock: Test MSE (mean and std. dev. over 5 different runs)}
    \label{tab:bharti_mf}
    \centering
    \footnotesize
    \begin{tabular}{ccccc}
        \hline
        (Input, Output) window & MLP & CNN & GRU-RNN & LSTM-RNN \\ \hline
        (30, 7)     & 0.0007121 (0.0000345) & 0.0008513 (0.0001565) & 0.0009969 (0.0001635) & 0.0010978 (0.0001433) \\ 
        (30, 14) & 0.0010981 (0.0000389) & 0.0011236 (0.0000643) & 0.0011692 (0.0002449) & 0.0011961 (0.0000745) \\ 
        (30, 21) & 0.0012744 (0.0000772) & 0.0012579 (0.0000285) & 0.0015933 (0.0001660) & 0.0016921 (0.0001325) \\ 
        (30, 28) & 0.0015675 (0.0001162) & 0.0015081 (0.0001651) & 0.0017918 (0.0001788) & 0.0021673 (0.0003877) \\ \hline
        (60, 7)     & 0.0010015 (0.0002596) & 0.0010944 (0.0002717) & 0.0010745 (0.0005075) & 0.0012880 (0.0002166) \\ 
        (60, 14) & 0.0011869 (0.0001420) & 0.0014233 (0.0003696) & 0.0014679 (0.0002400) & 0.0012717 (0.0001411) \\ 
        (60, 21) & 0.0013676 (0.0001464) & 0.0014766 (0.0000671) & 0.0019011 (0.0001783) & 0.0017968 (0.0007132) \\ 
        (60, 28) & 0.0016297 (0.0002142) & 0.0019439 (0.0001053) & 0.0023031 (0.0002640) & 0.0026234 (0.0006751) \\ \hline
        (90, 7)     & 0.0011117 (0.0002259) & 0.0012117 (0.0002445) & 0.0011537 (0.0001674) & 0.0016117 (0.0005630) \\ 
        (90, 14) & 0.0014819 (0.0004442) & 0.0017466 (0.0002888) & 0.0013579 (0.0001385) & 0.0014604 (0.0000723) \\ 
        (90, 21) & 0.0016159 (0.0002179) & 0.0019717 (0.0002747) & 0.0019266 (0.0002001) & 0.0018657 (0.0001663) \\ 
        (90, 28) & 0.0018643 (0.0003053) & 0.0020731 (0.0000764) & 0.0030601 (0.0011354) & 0.0022151 (0.0002840) \\ \hline
    \end{tabular}
\end{table*}

\begin{table*}[h]
    \caption{Multi-step forecast results for CIPLA stock: Test MSE (mean and std. dev. over 5 different runs)}
    \label{tab:cipla_mf}
    \centering
    \footnotesize
    \begin{tabular}{ccccc}
        \hline
        (Input, Output) window & MLP & CNN & GRU-RNN & LSTM-RNN \\ \hline
        (30, 7)  & 0.0004201 (0.0000362) & 0.0004331 (0.0000285) & 0.0008517 (0.0010172) & 0.0006759 (0.0002686) \\ 
        (30, 14) & 0.0004461 (0.0000768) & 0.0011685 (0.0005255) & 0.0005645 (0.0002275) & 0.0005727 (0.0003765) \\ 
        (30, 21) & 0.0005684 (0.0000595) & 0.0016531 (0.0006358) & 0.0006004 (0.0000952) & 0.0006103 (0.0001419) \\ 
        (30, 28) & 0.0008273 (0.0000740) & 0.0020585 (0.0002764) & 0.0006305 (0.0000400) & 0.0006017 (0.0000733) \\ \hline
        (60, 7)  & 0.0006738 (0.0002826) & 0.0005085 (0.0000557) & 0.0004554 (0.0001790) & 0.0006545 (0.0002807) \\ 
        (60, 14) & 0.0010456 (0.0002915) & 0.0007123 (0.0001254) & 0.0008674 (0.0005314) & 0.0007284 (0.0001981) \\ 
        (60, 21) & 0.0011300 (0.0002376) & 0.0007077 (0.0001503) & 0.0010405 (0.0002846) & 0.0008404 (0.0005486) \\ 
        (60, 28) & 0.0015181 (0.0003672) & 0.0013755 (0.0006623) & 0.0009845 (0.0003440) & 0.0008004 (0.0002936) \\ \hline
        (90, 7)  & 0.0011760 (0.0006906) & 0.0005961 (0.0002675) & 0.0005299 (0.0000354) & 0.0009868 (0.0010842) \\ 
        (90, 14) & 0.0026245 (0.0006215) & 0.0011966 (0.0008937) & 0.0015202 (0.0009906) & 0.0007429 (0.0001703) \\ 
        (90, 21) & 0.0018995 (0.0003939) & 0.0011213 (0.0003460) & 0.0011317 (0.0005927) & 0.0012397 (0.0004427) \\ 
        (90, 28) & 0.0039327 (0.0044482) & 0.0009193 (0.0000588) & 0.0012073 (0.0003054) & 0.0007912 (0.0000699) \\ \hline
    \end{tabular}
\end{table*}

\begin{table*}[h]
    \caption{Multi-step forecast results for HCLTECH stock: Test MSE (mean and std. dev. over 5 different runs)}
    \label{tab:hcl_mf}
    \centering
    \footnotesize
    \begin{tabular}{ccccc}
        \hline
        (Input, Output) window & MLP & CNN & GRU-RNN & LSTM-RNN \\ \hline
       (30, 7)  & 0.0031963 (0.0007954) & 0.0128769 (0.0019133) &    0.0051413 (0.0010541) &    0.0112670 (0.0021289) \\ 
       (30, 14) & 0.0026673 (0.0010671) & 0.0139527 (0.0055610) &    0.0123741 (0.0021109) &    0.0095371 (0.0062635) \\ 
       (30, 21) & 0.0033571 (0.0017455) & 0.0184534 (0.0016407) &    0.0118404 (0.0023245) &    0.0107927 (0.0043291) \\ 
       (30, 28) & 0.0038904 (0.0016345) & 0.0151611 (0.0020196) &    0.0244928 (0.0057851) &    0.0104605 (0.0035029) \\ \hline
       (60, 7)  & 0.0021283 (0.0014088) & 0.0109632 (0.0021675) &    0.0077522 (0.0041052) &    0.0143966 (0.0060136) \\ 
       (60, 14) & 0.0038363 (0.0016056) & 0.0213113 (0.0066804) &    0.0132269 (0.0031572) &    0.0094325 (0.0038677) \\ 
       (60, 21) & 0.0038672 (0.0013679) & 0.0162212 (0.0083456) &    0.0235211 (0.0040127) &    0.0081314 (0.0040108) \\ 
       (60, 28) & 0.0054724 (0.0022313) & 0.0196803 (0.0112306) &    0.0319193 (0.0048348) &    0.0123502 (0.0060779) \\ \hline
       (90, 7)  & 0.0058540 (0.0025836) & 0.0105892 (0.0029056) &    0.0008600 (0.0007130) &    0.0051682 (0.0030051) \\ 
       (90, 14) & 0.0071620 (0.0033313) & 0.0309257 (0.0070107) &    0.0035058 (0.0028799) &    0.0050845 (0.0028523) \\ 
       (90, 21) & 0.0077732 (0.0048928) & 0.0241231 (0.0102371) &    0.0086925 (0.0014761) &    0.0074075 (0.0057231) \\ 
       (90, 28) & 0.0075716 (0.0064593) & 0.0270582 (0.0119271) &    0.0082987 (0.0017760) &    0.0055403 (0.0024363) \\ \hline
    \end{tabular}
\end{table*}

\begin{table*}[h]
    \caption{Multi-step forecast results for HDFC stock: Test MSE (mean and std. dev. over 5 different runs)}
    \label{tab:hdfc_mf}
    \centering
    \footnotesize
    \begin{tabular}{ccccc}
        \hline
        (Input, Output) window & MLP & CNN & GRU-RNN & LSTM-RNN \\ \hline
        (30, 7)  & 0.0014993 (0.0001856) & 0.0022640 (0.0008543) & 0.0004043 (0.0002280) & 0.0016212 (0.0010235) \\ 
        (30, 14) & 0.0029107 (0.0006320) & 0.0035546 (0.0007540) & 0.0008814 (0.0003569) & 0.0008856 (0.0004405) \\ 
        (30, 21) & 0.0046339 (0.0009452) & 0.0056335 (0.0004270) & 0.0008844 (0.0003937) & 0.0006469 (0.0001056) \\ 
        (30, 28) & 0.0056033 (0.0006032) & 0.0082196 (0.0016140) & 0.0017710 (0.0006404) & 0.0011004 (0.0000466) \\ \hline
        (60, 7)  & 0.0024260 (0.0005739) & 0.0034951 (0.0027383) & 0.0012631 (0.0008066) & 0.0017102 (0.0016013) \\ 
        (60, 14) & 0.0034203 (0.0012190) & 0.0047431 (0.0005306) & 0.0012097 (0.0007921) & 0.0013114 (0.0007453) \\ 
        (60, 21) & 0.0066358 (0.0007448) & 0.0081363 (0.0028405) & 0.0008241 (0.0000640) & 0.0014144 (0.0011814) \\ 
        (60, 28) & 0.0065224 (0.0016475) & 0.0119449 (0.0034957) & 0.0013372 (0.0001863) & 0.0012217 (0.0001187) \\ \hline
        (90, 7)  & 0.0023300 (0.0012620) & 0.0090839 (0.0042322) & 0.0033794 (0.0024053) & 0.0021608 (0.0024645) \\ 
        (90, 14) & 0.0054503 (0.0017082) & 0.0099783 (0.0047804) & 0.0014637 (0.0006768) & 0.0012345 (0.0006202) \\ 
        (90, 21) & 0.0077797 (0.0016669) & 0.0115749 (0.0095457) & 0.0010936 (0.0005853) & 0.0009897 (0.0001870) \\ 
        (90, 28) & 0.0095018 (0.0031604) & 0.0112411 (0.0038547) & 0.0017048 (0.0006377) & 0.0011751 (0.0004205) \\ \hline
    \end{tabular}
\end{table*}

\begin{table*}[h]
    \caption{Multi-step forecast results for INFY stock: Test MSE (mean and std. dev. over 5 different runs)}
    \label{tab:infy_mf}
    \centering
    \footnotesize
    \begin{tabular}{ccccc}
        \hline
        (Input, Output) window & MLP & CNN & GRU-RNN & LSTM-RNN \\ \hline
        (30, 7)  & 0.0008039 (0.0002816) & 0.0005298 (0.0000788) & 0.0004394 (0.0001770) & 0.0003833 (0.0000810) \\ 
        (30, 14) & 0.0010837 (0.0006352) & 0.0007653 (0.0002916) & 0.0005086 (0.0001160) & 0.0005351 (0.0000950) \\ 
        (30, 21) & 0.0011262 (0.0002798) & 0.0010261 (0.0001624) & 0.0010603 (0.0002050) & 0.0011424 (0.0001741) \\ 
        (30, 28) & 0.0010185 (0.0002701) & 0.0010253 (0.0002893) & 0.0007198 (0.0001526) & 0.0007297 (0.0000437) \\ \hline
        (60, 7)  & 0.0012397 (0.0001411) & 0.0004898 (0.0002666) & 0.0003308 (0.0001858) & 0.0003605 (0.0000608) \\ 
        (60, 14) & 0.0013622 (0.0007259) & 0.0006215 (0.0001482) & 0.0006564 (0.0001803) & 0.0010193 (0.0004008) \\ 
        (60, 21) & 0.0013731 (0.0007609) & 0.0008720 (0.0001914) & 0.0007246 (0.0000544) & 0.0015434 (0.0000978) \\ 
        (60, 28) & 0.0011040 (0.0000702) & 0.0012745 (0.0002797) & 0.0008392 (0.0001180) & 0.0009567 (0.0002125) \\ \hline
        (90, 7)  & 0.0011997 (0.0006184) & 0.0005635 (0.0005604) & 0.0003365 (0.0000540) & 0.0006026 (0.0002255) \\ 
        (90, 14) & 0.0018267 (0.0007732) & 0.0010913 (0.0006205) & 0.0006243 (0.0000194) & 0.0007335 (0.0000828) \\ 
        (90, 21) & 0.0013901 (0.0009740) & 0.0009851 (0.0002199) & 0.0009886 (0.0000533) & 0.0010992 (0.0000903) \\ 
        (90, 28) & 0.0019559 (0.0012953) & 0.0014412 (0.0003111) & 0.0008899 (0.0000883) & 0.0010402 (0.0001649) \\ \hline
    \end{tabular}
\end{table*}

\begin{table*}[h]
    \caption{Multi-step forecast results for JSWSTEEL stock: Test MSE (mean and std. dev. over 5 different runs)}
    \label{tab:jsw_mf}
    \centering
    \footnotesize
    \begin{tabular}{ccccc}
        \hline
        (Input, Output) window & MLP & CNN & GRU-RNN & LSTM-RNN \\ \hline
        (30, 7)  & 0.0008159 (0.0002204) & 0.0015366 (0.0010747) & 0.0007248 (0.0004859) & 0.0010302 (0.0008532) \\ 
        (30, 14) & 0.0012793 (0.0006275) & 0.0008200 (0.0002332) & 0.0011439 (0.0003228) & 0.0010638 (0.0005344) \\ 
        (30, 21) & 0.0009010 (0.0002266) & 0.0008294 (0.0002427) & 0.0008718 (0.0003286) & 0.0007978 (0.0001207) \\
        (30, 28) & 0.0014537 (0.0003449) & 0.0008152 (0.0001385) & 0.0014504 (0.0002590) & 0.0015659 (0.0009209) \\ \hline
        (60, 7)  & 0.0006371 (0.0002588) & 0.0035135 (0.0014058) & 0.0010019 (0.0007528) & 0.0005054 (0.0002913) \\
        (60, 14) & 0.0015956 (0.0002350) & 0.0038904 (0.0019989) & 0.0008280 (0.0005101) & 0.0010641 (0.0009801) \\ 
        (60, 21) & 0.0015652 (0.0003992) & 0.0023567 (0.0006488) & 0.0007918 (0.0002060) & 0.0009342 (0.0006045) \\ 
        (60, 28) & 0.0022744 (0.0012208) & 0.0024875 (0.0011985) & 0.0010372 (0.0002349) & 0.0011837 (0.0004319) \\ \hline
        (90, 7)  & 0.0011166 (0.0004300) & 0.0035503 (0.0022253) & 0.0004170 (0.0002284) & 0.0005311 (0.0001889) \\ 
        (90, 14) & 0.0016961 (0.0004295) & 0.0049484 (0.0025952) & 0.0005950 (0.0003462) & 0.0006270 (0.0002636) \\ 
        (90, 21) & 0.0019113 (0.0002832) & 0.0041109 (0.0006370) & 0.0007005 (0.0001861) & 0.0006317 (0.0001386) \\ 
        (90, 28) & 0.0023677 (0.0004270) & 0.0038738 (0.0010972) & 0.0009316 (0.0002275) & 0.0015127 (0.0005895) \\ \hline
    \end{tabular}
\end{table*}

\begin{table*}[h]
    \caption{Multi-step forecast results for MARUTI stock: Test MSE (mean and std. dev. over 5 different runs)}
    \label{tab:maruti_mf}
    \centering
    \footnotesize
    \begin{tabular}{ccccc}
        \hline
        (Input, Output) window & MLP & CNN & GRU-RNN & LSTM-RNN \\ \hline
        (30, 7)  & 0.0036446 (0.0014446) & 0.0053164 (0.0022883) & 0.1113903 (0.0217113) & 0.0341339 (0.0091429) \\ 
        (30, 14) & 0.0044910 (0.0008551) & 0.0064419 (0.0009121) & 0.0659209 (0.0611601) & 0.0559117 (0.0358284) \\ 
        (30, 21) & 0.0084389 (0.0010078) & 0.0082142 (0.0006503) & 0.0914546 (0.0711125) & 0.0430067 (0.0210945) \\ 
        (30, 28) & 0.0124981 (0.0037026) & 0.0187498 (0.0043378) & 0.0661392 (0.0479253) & 0.0643477 (0.0320369) \\ \hline
        (60, 7)  & 0.0084413 (0.0036564) & 0.0144930 (0.0102842) & 0.0265717 (0.0219347) & 0.0376275 (0.0094258) \\ 
        (60, 14) & 0.0085299 (0.0031797) & 0.0139296 (0.0035586) & 0.1302003 (0.1588328) & 0.0374050 (0.0193156) \\ 
        (60, 21) & 0.0123392 (0.0037315) & 0.0308576 (0.0071213) & 0.0778246 (0.0261484) & 0.0631713 (0.0226513) \\ 
        (60, 28) & 0.0151471 (0.0025340) & 0.0617911 (0.0132642) & 0.0619888 (0.0188969) & 0.0508528 (0.0106388) \\ \hline
        (90, 7)  & 0.0121688 (0.0057908) & 0.0139867 (0.0067465) & 0.0188332 (0.0087000) & 0.0335358 (0.0163816) \\ 
        (90, 14) & 0.0152392 (0.0052767) & 0.0332158 (0.0108789) & 0.0510381 (0.0252156) & 0.0399558 (0.0212659) \\ 
        (90, 21) & 0.0204686 (0.0027830) & 0.0607542 (0.0224152) & 0.0411452 (0.0110080) & 0.0650783 (0.0372264) \\ 
        (90, 28) & 0.0206555 (0.0054890) & 0.0627885 (0.0128993) & 0.0401244 (0.0165746) & 0.0865678 (0.0323832) \\ \hline
    \end{tabular}
\end{table*}

\begin{table*}[h]
    \caption{Multi-step forecast results for ULTRACEMCO stock: Test MSE (mean and std. dev. over 5 different runs)}
    \label{tab:ultra_mf}
    \centering
    \footnotesize
    \begin{tabular}{ccccc}
        \hline
        (Input, Output) window & MLP & CNN & GRU-RNN & LSTM-RNN \\ \hline
        (30, 7)  & 0.0020023 (0.0002984) & 0.0026930 (0.0005455) & 0.0033386 (0.0020864) & 0.0017219 (0.0006294) \\ 
        (30, 14) & 0.0024367 (0.0001538) & 0.0033537 (0.0001399) & 0.0044677 (0.0010579) & 0.0019778 (0.0005814) \\
        (30, 21) & 0.0033515 (0.0001442) & 0.0037870 (0.0001854) & 0.0056589 (0.0037724) & 0.0027700 (0.0002355) \\ 
        (30, 28) & 0.0038125 (0.0004258) & 0.0039539 (0.0000881) & 0.0045104 (0.0012098) & 0.0030197 (0.0002363) \\ \hline
        (60, 7)  & 0.0024746 (0.0002628) & 0.0042345 (0.0016226) & 0.0031997 (0.0014948) & 0.0017283 (0.0005668) \\
        (60, 14) & 0.0034616 (0.0005681) & 0.0046157 (0.0013290) & 0.0048650 (0.0016006) & 0.0022667 (0.0006604) \\ 
        (60, 21) & 0.0034066 (0.0003507) & 0.0045258 (0.0005243) & 0.0031052 (0.0005199) & 0.0023807 (0.0003801) \\ 
        (60, 28) & 0.0040590 (0.0002551) & 0.0049600 (0.0005849) & 0.0098274 (0.0039146) & 0.0030861 (0.0002780) \\ \hline
        (90, 7)  & 0.0040636 (0.0011404) & 0.0090143 (0.0035609) & 0.0025351 (0.0015730) & 0.0020238 (0.0007153) \\ 
        (90, 14) & 0.0044154 (0.0006285) & 0.0063803 (0.0038330) & 0.0079744 (0.0019826) & 0.0020690 (0.0004533) \\ 
        (90, 21) & 0.0047966 (0.0007563) & 0.0052869 (0.0008835) & 0.0075306 (0.0049383) & 0.0025133 (0.0002801) \\ 
        (90, 28) & 0.0059788 (0.0006318) & 0.0066679 (0.0017524) & 0.0163860 (0.0097798) & 0.0029608 (0.0003193) \\ \hline
    \end{tabular}
\end{table*}

\begin{figure*}[h]
    \centering
    \subfloat[$inp_w = 30, output_w = 7$]{\includegraphics[width = 3in, height = 1.6in]{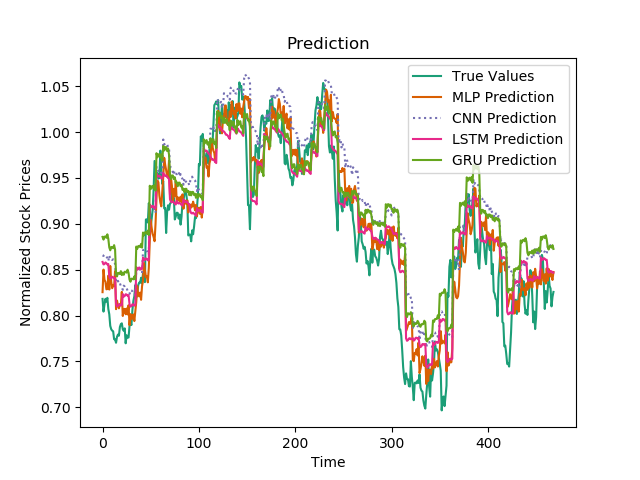}} 
    \subfloat[$inp_w = 30, output_w = 28$]{\includegraphics[width = 3in, height = 1.6in]{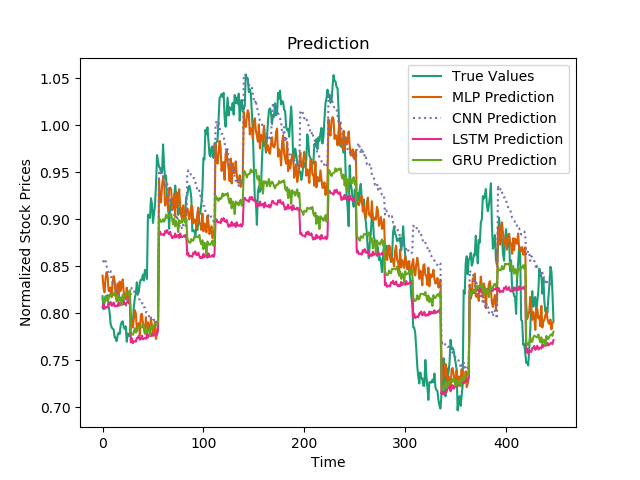}}\\
    \subfloat[$inp_w = 60, output_w = 7$]{\includegraphics[width = 3in, height = 1.6in]{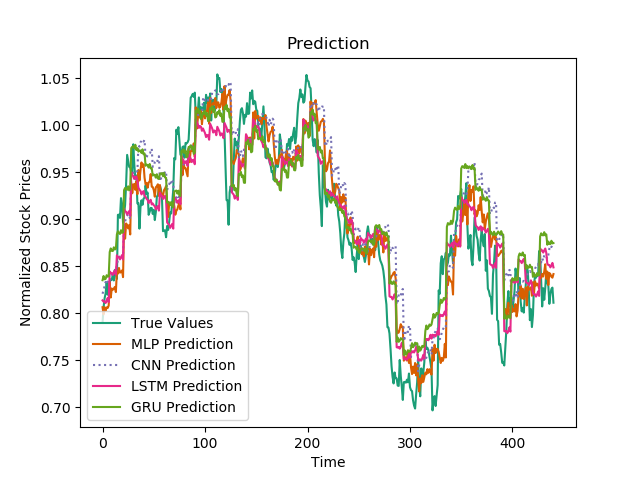}} 
    \subfloat[$inp_w = 60, output_w = 28$]{\includegraphics[width = 3in, height = 1.6in]{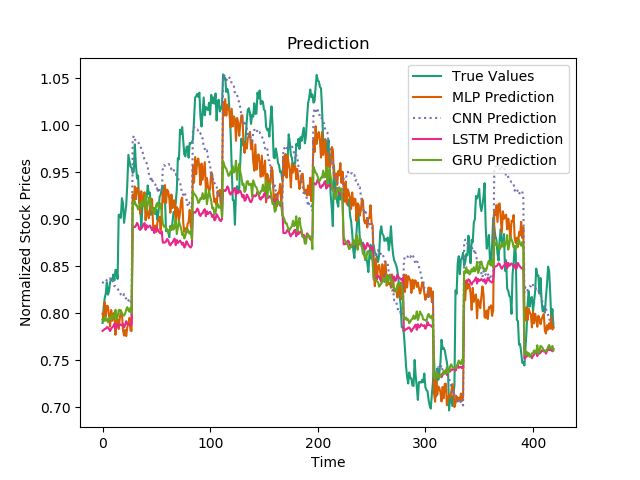}}\\
    \subfloat[$inp_w = 90, output_w = 7$]{\includegraphics[width = 3in, height = 1.6in]{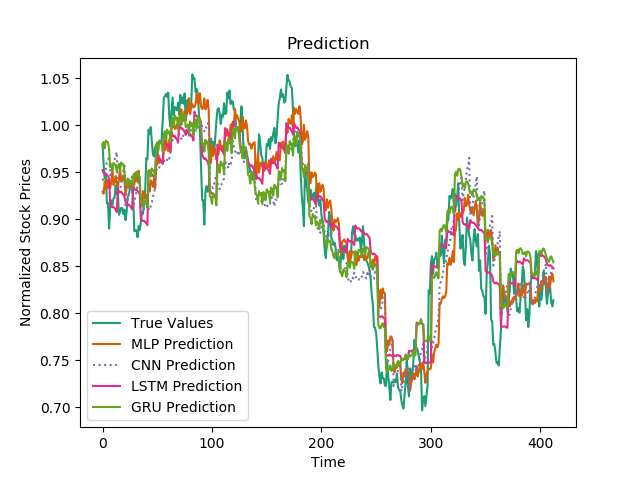}} 
    \subfloat[$inp_w = 90, output_w = 28$]{\includegraphics[width = 3in, height = 1.6in]{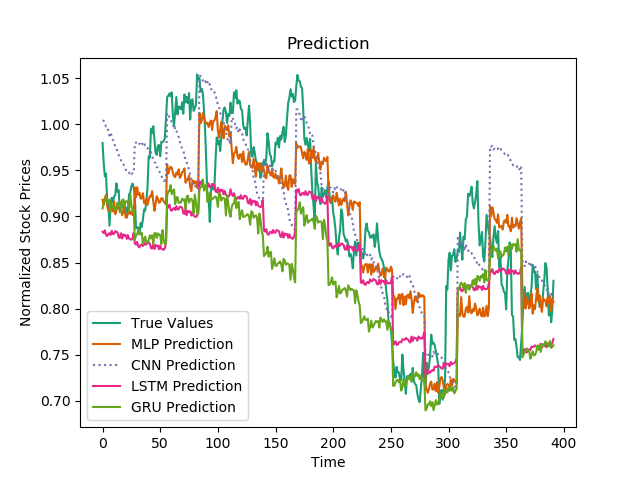}}\\
    \caption{Multi-step forecasting plots for ACC dataset for each deep model against the true prices for different window size combinations($w$). Due to reason of space, We are unable to provide the plots for the output window sizes: 14 and 21.}
\end{figure*}

\end{document}